\documentclass[pdflatex,sn-basic]{sn-jnl}

\usepackage{amsmath}
\usepackage{caption}
\usepackage{subcaption}
\DeclareMathOperator*{\argmin}{argmin}

\theoremstyle{thmstyleone}%
\theoremstyle{thmstyletwo}%

\theoremstyle{thmstylethree}%

\begin{document}

\title[Studying Therapy Effects in Silico]{Studying Therapy Effects and Disease Outcomes in Silico using Artificial Counterfactual Tissue Samples}

\author*[1]{\fnm{Martin} \sur{Paulikat}}

\author[2]{\fnm{Christian M.} \sur{Schürch}}
\equalcont{These authors contributed equally to this work and share last authorship.}

\author[1]{\fnm{Christian F.} \sur{Baumgartner}}
\equalcont{These authors contributed equally to this work and share last authorship.}

\affil*[1]{Cluster of Excellence - Machine Learning for Science, University of Tübingen, Tübingen, Germany}

\affil[2]{Department of Pathology and Neuropathology, University Hospital and Comprehensive Cancer Center Tübingen, Tübingen, Germany}


\abstract{Understanding the interactions of different cell types inside the immune tumor microenvironment (iTME) is crucial for the development of immunotherapy treatments as well as for predicting their outcomes. Highly multiplexed tissue imaging (HMTI) technologies offer a tool which can capture cell properties of tissue samples by measuring expression of various proteins and storing them in separate image channels. HMTI technologies can be used to gain insights into the iTME and in particular how the iTME differs for different patient outcome groups of interest (e.g., treatment responders vs. non-responders). Understanding the systematic differences in the iTME of different patient outcome groups is crucial for developing better treatments and personalising existing treatments. However, such analyses are inherently limited by the fact that any two tissue samples vary due to a large number of factors unrelated to the outcome. Here, we present CF-HistoGAN, a machine learning framework that employs generative adversarial networks (GANs) to create artificial counterfactual tissue samples that resemble the original tissue samples as closely as possible but capture the characteristics of a different patient outcome group. Specifically, we learn to "translate" HMTI samples from one patient group to create artificial paired samples. We show that this approach allows to directly study the effects of different patient outcomes on the iTMEs of individual tissue samples. We demonstrate that CF-HistoGAN can be employed as an explorative tool for understanding iTME effects on the pixel level. Moreover, we show that our method can be used to identify statistically significant differences in the expression of different proteins between patient groups with greater sensitivity compared to conventional approaches.}




\maketitle

\section{Introduction}

Cancer is a leading global disease with almost 20 million new cases annually, resulting in around 10 million deaths~\citep{sung_global_2021}. Established cancers do not only consist of tumor cells; they represent complex ecosystems that incorporate various other cell types including vascular endothelial cells, fibroblastic stromal cells and a wide variety of immune cell types, collectively termed the “immune tumor microenvironment” (iTME)~\citep{binnewies_understanding_2018}. The iTME is crucial for the initiation and progression of cancers by inducing a tumor-promoting chronic inflammation and an immunosuppressive milieu that mitigates antitumoral immunity~\citep{hanahan2011hallmarks}. Therefore, the iTME is the main site of interactions between tumor and immune cells, and its composition substantially affects disease outcomes~\citep{schreiber2011cancer,junttila2013influence}. In recent years, the development of immunotherapies, especially immune checkpoint blockers (ICBs), has created a paradigm shift in cancer therapy. Treatments with ICBs, e.g., antibodies against programmed cell death-1 (PD-1), its ligand PD-L1, or cytotoxic T cell antigen-4 (CTLA-4), have resulted in long-lasting remissions in a variety of advanced, even metastatic tumors~\citep{melero_evolving_2015}. However, more than 70\% of patients treated with ICBs do not respond to this therapy~\citep{topalian_mechanism-driven_2016}. In addition, ICB treatments can result in devastating adverse effects~\citep{michot_immune-related_2016}, and pose enormous economic burdens on healthcare systems~\citep{ghate_economic_2018}. 

The spatial cell arrangement of the iTME can give us insights into response to ICB treatments and patient survival. CODEX (co-detection by indexing) is a highly multiplexed tissue imaging (HMTI) technology that uses DNA-conjugated antibodies and iterative hybridization and stripping of fluorophore-tagged DNA probes to achieve high-parameter immunofluorescence imaging of fresh-frozen or paraffin-embedded tissue~\citep{black2021codex,kennedy2021highly}. The resulting multi-channel images can be used to gain an understanding of differences in the spatial cell arrangement of the iTME for different groups of interest by comparing their respective CODEX images. In prior work, we studied the iTME of colorectal cancer (CRC) patients with good and poor survival and identified coordinated “cellular neighborhoods”, regions of the iTME with specific properties, whose functions were associated with patient outcomes~\citep{schurch_coordinated_2020}. Moreover, using CODEX on cutaneous T cell lymphoma (CTCL) samples, we could demonstrate that the physical distances between certain immune cell types and tumor cells in the iTME could predict whether a patient responded to immunotherapy or not~\citep{phillips_immune_2021}. In addition, we used this imaging technology to study therapy responses to tumor necrosis factor inhibitors in inflammatory bowel disease~\citep{mayer2023tissue}, and to dissect immune cell rearrangements in the iTME of glioblastoma ex vivo 3D bioreactor models treated with different immunotherapies targeting the innate and adaptive immune system~\citep{shekarian2022immunotherapy}.

However, CODEX images are challenging to interpret, due the high dimensionality of the data, the presence of noise and artifacts in the images, and the complexity of the tissue structures being imaged. This makes applying automated tools such as segmentation and cell type detection challenging. Additionally, the use of multiple channels to capture different parameters of the cells can make it difficult to interpret all of the information simultaneously and visually identify potential systematic differences between patient groups. Lastly, tissue samples appear vastly different on a pixel level due to the unique cell arrangements of each sample. This prevents direct visual comparison between regions of interest for different patient outcome groups. 

In this paper, we propose a histopathological image analysis technique based on machine learning which enables generating counterfactual tissue samples. That is, it allows “translating” a sample of one patient outcome group (e.g., treatment non-responders), to another outcome group (e.g., treatment responders), while preserving the overall cell arrangement of the sample. This allows direct visual comparison of a tissue sample to the ``what if?'' scenario of this sample corresponding to a different outcome group. The method is based on generative adversarial networks (GANs) and operates directly on the raw pixel level, thus circumventing the need for segmentation and downstream analysis algorithms. The generated counterfactuals could be a valuable tool for obtaining biological hypotheses about disease mechanisms. For example, by examining spatial areas that are altered in each channel of the CODEX images, systematic differences between outcome groups in protein expression or cell arrangement may become apparent. Moreover, by generating a counterfactual sample for each tissue sample, we can statistically compare all samples to their respective artificial controls. This allows identifying systematic differences in staining intensities between two patient outcome groups with high sensitivity. 

Counterfactual analysis of medical images has recently gained attention with the advent of powerful generative modelling techniques based on neural networks. \citet{pawlowski2020deep} formulate deep structural causal models which allow them to obtain counterfactuals by performing interventions on the causal graph. The authors explore this approach on neuroimaging data. In an extension of this work, \citet{rasal2022deep} explore counterfactuals on 3D shape models of brain structures. Taking a different approach, we have previously proposed directly transforming the apparent disease status of patients with Alzheimer's disease using GANs~\citep{baumgartner_visual_2018}. By analysing the difference between the original images with their generated counterfactual counterparts, we were able to create subject specific disease effect maps. The work most closely related to the present paper is the Cell-Cell Interaction GAN (CCIGAN) proposed by \citet{li_counterfactual_2020}. The authors train a GAN to generate realistic HMTI samples from cell segmentation masks. By inputting altered cell segmentation maps into this network the authors can explore counterfactual cell arrangement scenarios. In contrast to this work, here, we study the effect of different patient outcome groups. The proposed work is, to our knowledge, the first to explore the hypothetical effects of patient group membership on the raw pixels of HMTI histopathology data. 

Our contributions are as follows:
\begin{itemize}
    \item We extend our previously proposed method for visual attribution using generative adversarial networks~\citep{baumgartner_visual_2018} to counterfactual generation of CODEX images. We coin this new approach the counterfactual histopathology GAN (CF-HistoGAN). 
    \item We demonstrate that our proposed CF-HistoGAN can successfully transform CODEX images between the groups of good and poor survival in CRC, and treatment responders and non-responders in CTCL. 
    \item We further demonstrate that, by generating artificial control samples, our proposed method can identify biomarkers that are correlated with the studied patient groups for both of the above cancers with greater sensitivity compared to conventional unpaired analysis of samples. 
\end{itemize}

\section{Material and Equipment}
\label{sec:material-and-equipment}

We evaluate the proposed CF-HistoGAN approach on two cancer datasets which we have studied in the past. A CRC dataset, where the clinical variable of interest is poor versus good survival~\citep{schurch_coordinated_2020}, and a CTCL dataset where we study treatment responders versus treatment non-responders~\citep{phillips_immune_2021}. Both datasets were collected using the CODEX HMTI technique.

\subsection{Colorectal cancer (CRC) dataset}

The CRC dataset is composed of highly multiplexed images from tissue microarray cores taken from 35 advanced-stage CRC patients with good and poor outcomes~\citep{schurch_coordinated_2020}. All TMA cores are 0.6\,mm in diameter and 4\,µm thick. 17 CRC patients belong to the group with a Crohn’s-like reaction (CLR), a strong adaptive immune response with many tertiary lymphoid structures (follicles) which correlates with good survival. The remaining 18 patients are in the diffuse inflammatory infiltration (DII) group, which correlates with poor survival. The CRC dataset contains 140 image files (4 images per patient), 68 image files for the CLR group, and 72 for the DII group. Each image file contains 58 unique protein expresion channels and 3 channels for the hematoxylin and eosin (H\&E) stain. Each image has a dimensionality of 1920x1440 pixels.

\subsection{Cutaneous T cell lymphoma (CTCL) dataset}
The CTCL dataset is composed of TMA cores taken from patients who suffer from CTCL, a cancer of the lymphatic system manifesting in the skin~\citep{phillips_immune_2021}. CTCL patients were treated with pembrolizumab, an anti-PD-1 antibody, in the framework of a clinical trial, and skin biopsies were taken before and after treatment~\citep{khodadoust2020pembrolizumab}. All TMA cores are 0.6\,mm in diameter and 4\,µm thick. Like the CRC dataset, the raw images in the CTCL dataset were processed using the CODEX Toolkit uploader~\citep{goltsev_deep_2018}. The dataset consists of samples from 14 patients, of which 7 were treatment responders and 7 were non-responders. Overall, the dataset contains 70 image files, with 32 images belonging to the responders group and 38 images belonging to the non-responders group. Each image file contains 59 unique protein expression channels and 3 channels for the H\&E stain. As before, each image has a dimensionality of 1920x1440 pixels.

\section{Methods}

In the following, we describe our machine learning approach for generating counterfactual CODEX tissue samples as well as the preprocessing, hyper-parameter tuning methods, and evaluation methods. 

\begin{figure}
    \centering
    \includegraphics[width=0.99\textwidth]{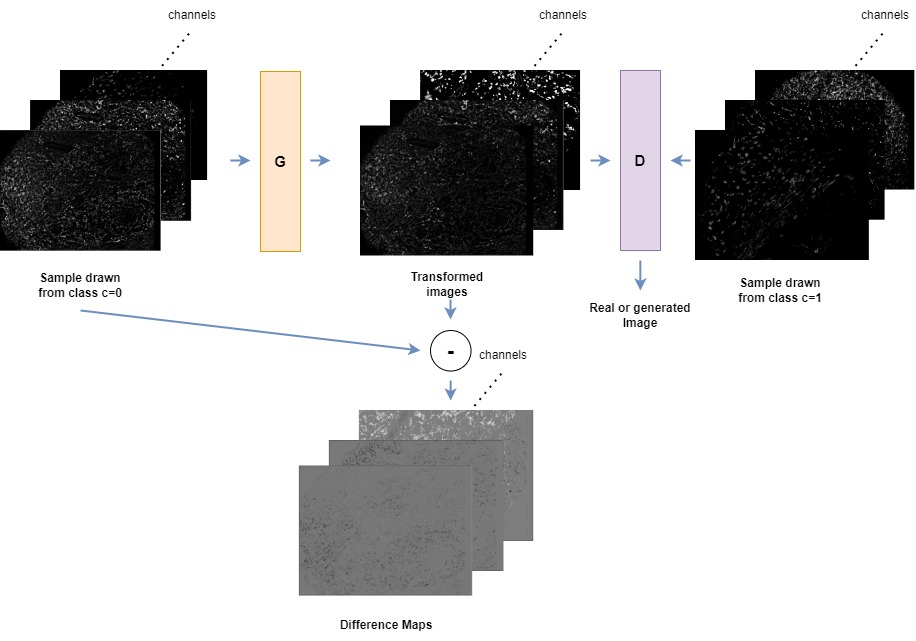}
    \caption{CF-HistoGAN overview. The generator function $G$ transforms samples drawn from one patient outcome group $c=0$ to resemble samples of another outcome group $c=1$. The discriminator function $D$ simultaneously learns to distinguish real samples drawn from $c=1$ from artificially generated ones. Both $G$ and $D$ can process and multi-channel images. The final difference maps are computed by simply subtracting the original sample from the transformed image and contain visually interpretable map of which channels and areas of the input image would look differently if the image came from a different patient outcome group.}
    \label{fig:overview}
\end{figure}

\subsection{Counterfactual Histopathology GAN (CF-HistoGAN)}

Our method aims to generate tissue samples that differ only in the patient outcome group (e.g. poor survival versus good survival, or treatment responders versus nonresponders) but, in all other respects, remain as similar to the original tissue sample as possible. To achieve this, we draw upon our previously published GAN-based approach for counterfactual visualisation of subject specific effects of Alzheimer's disease~\citep{baumgartner_visual_2018}. In the present work, we apply a very similar methodology to CODEX images. Specifically, we assume we have image data from two classes $c \in \{0, 1\}$  (e.g., treatment responders and nonresponders). We formulate the problem as learning a generator network $G_\theta$, parametrised by parameters $\theta$, that transforms images from class $c=0$, to class $c=1$. In order to facilitate the problem we pose it as the estimation of the \textit{residual} which needs to be added to the input image in order to change it into the other class. To this end, we introduce a residual generator function $M_\theta$, and write the counterfactual generation problem as
\begin{equation}
    y^{(1)} =  G_\theta(x^{(0)}) = \sigma \Bigl(x^{(0)} + M_\theta(x^{(0)} )\Bigr),
\end{equation}
where $x^{(0)}$ is a real multi-channel image from $c=0$, $y^{(1)}$ is a generated multi-channel image from $c=1$, and $\sigma$ is the sigmoid function ensuring that the output pixels for each channel fall in the range $(0,1)$, which matches the range of the input data. We found that using the sigmoid functions on the outputs improved the training behavior and quality of the generated images. The residual generator function $M_\theta(\cdot)$ is parameterized as a UNet-like neural network. An overview of the proposed framework is shown in Fig.~\ref{fig:overview}.

The main challenge in our approach is learning the function $G_\theta(\cdot)$ without access to paired samples from both patient outcome groups $c=0$ and $c=1$. To address this, we use an approach based on GANs as in our earlier work~\citep{baumgartner_visual_2018}. The goal is to make the generated image $y^{(1)}$ indistinguishable from real images in class $c=1$. To achieve this, we train an additional discriminator network $D_\varphi(\cdot)$, parametrised by parameters $\varphi$, to distinguish generated images $y^{(1)}$ from real images $x^{(1)}$. The generator's objective is to create output images that the discriminator can no longer distinguish from real images. In order to encourage the generator network to change \textit{only} the patient outcome group $c$ but leave the sample otherwise unchanged, we place a sparsity promoting $L_1$ constraint on the difference between the input image and the generated image.

Taking the approach used by the original Wasserstein GAN~\citep{arjovsky_wasserstein_2017} and our earlier work~\cite{baumgartner_visual_2018}, this leads to the following min-max objective: 
\begin{multline}
\label{eq:wgan}
  M^* = \argmin\limits_{\theta} \max\limits_{\varphi \ : \ D\in\mathcal{D}}  \mathbb{E}_{x^{(1)}\sim p(x \mid c=1)}\left[D(x^{(1)})\right] \\ 
  - \mathbb{E}_{x^{(0)} \sim p(x \mid c=0)}\left[G_\theta(x^{(0)}) + \lambda \vert\vert G_\theta(x^{(0)}) - x^{(0)} \vert\vert_1\right] .
\end{multline}
The constraint $D\in\mathcal{D}$ ensures that $D$ comes from the class 1-Lipschitz functions $\mathcal{D}$ as is required by the Wasserstein GAN theory~\cite{arjovsky_wasserstein_2017}. We enforce this constraint using the gradient penalty term introduced by \citet{gulrajani_improved_2017} which is omitted in Equation~\ref{eq:wgan} for brevity. The hyperparameter $\lambda$ controls the importance of the $L_1$ regularisation term, i.e. how closely the output image needs to resemble the input image. 

In contrast to the work by \citet{baumgartner_visual_2018}, the large size of CODEX images makes it impractical to directly optimize the objective function described above for entire image samples, as it exceeds the memory capacity of most current GPUs. To overcome this issue, we decompose the problem into patch-wise image translation. Specifically, we divide each CODEX image into overlapping square patches of fixed size and stride, and train the model on these patches. The stride relative to the patch size defines the overlap of the patches and indirectly controls how many training patches we can generate. Smaller strides lead to larger overlap between patches and lead to a larger training set. This can be considered a mechanism for data augmentation and proved crucial to generate sufficient data to train our proposed method. The prediction for an entire CODEX image can be obtained by stitching together the patch-wise predictions. This process can be parallelized on the GPU, allowing us to generate counterfactual CODEX images in approximately 10 seconds. As we will explain later, in this work, we use an ensemble of 9 networks which raises the prediction time to 90 seconds when using a single GPU.

\subsection{Network Architecutre}

\begin{figure}
     \centering
     \begin{subfigure}[b]{0.45\textwidth}
         \centering
         \includegraphics[width=\textwidth]{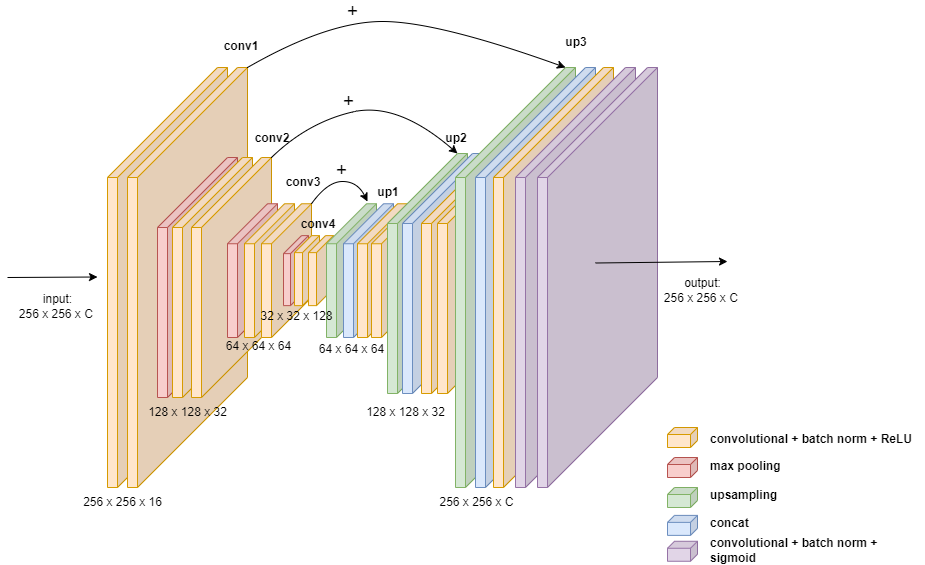}
         \caption{Generator}
         \label{fig:generator}
     \end{subfigure}
     \hfill
     \begin{subfigure}[b]{0.45\textwidth}
         \centering
         \includegraphics[width=\textwidth]{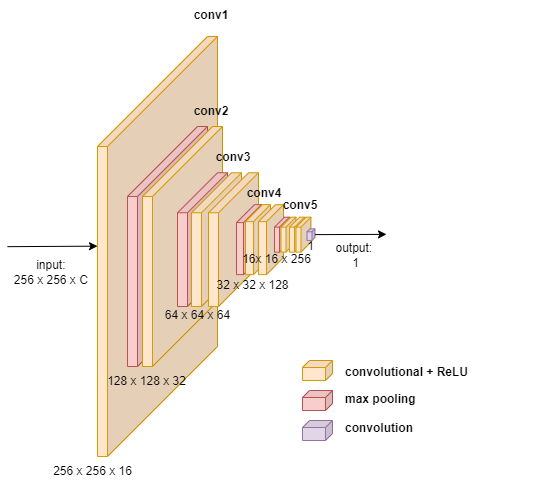}
         \caption{Discriminator}
         \label{fig:discriminator}
     \end{subfigure}
        \caption{Schematics of the network architectures for the generator network $G$ (a), and the discriminator network $D$. The parameter $C$ refers to the number of input channels, i.e. the number of biomarkers acquired in a CODEX image.}
        \label{fig:architecture}
\end{figure}

The network architectures for the generator and discriminator networks $G_\theta$ and $D_\varphi$ were loosely based on the architectures used in~\citep{baumgartner_visual_2018}. In contrast to that work, we do not work on gray-scale 3D images but rather on 2D patches of large multi-channel histopathology images. The architecture for both networks is summarised in Figure~\ref{fig:architecture}.

\subsection{Training}
\label{sec:training}

The channels of each CODEX image were individually normalized to a range from 0 to 1, and then divided into square patches with a specified pixel height and width $p$. These patches had a certain overlap, determined by the pixel stride $s$. The images were then scaled down by a factor $d$ using nearest neighbor interpolation.

For each dataset, we trained networks to transform images in both directions: for the CRC dataset, we trained networks to transform images with poor outcomes (DII) into images with good outcomes (CLR) and vice versa, and for the CTCL dataset, we trained a network to transform images of treatment responders into images of treatment nonresponders and vice versa. 

Each of those networks was trained using all images from the respective dataset. We did not perform a training/test/validation split as is usual for predictive methods. Note that in contrast to predictive tasks our objective here is not to train a network that generalizes to previously unseen test cases. Rather, we use our proposed CF-HistoGAN to explore relationships between two groups of interest within the training set. Thus, we do not need a test set to assess generalization performance. However, in order to monitor the training process, we used one image as validation to control for overfitting.

We trained the networks for a maximum of 500 epochs, which typically took 28 hours on a NVIDIA V100 GPU. During training, we observed significant fluctuations in the quality of the generated images from epoch to epoch. To ensure the reliability of our results and avoid the need for complicated model selection, we selected 9 training checkpoints evenly spaced between epochs 300 and 500 for the CRC data set, and between epochs 200 and 400 for the CTCL data set. Those ranges were empirically chosen based on our initial observations. The CTCL dataset required fewer epochs because it had fewer data points and therefore converged faster. To obtain a single prediction, we created an ensemble of 9 networks initialized with these intermediate checkpoints. 

\subsection{Hyper-parameter selection}

Our method relies on a number of hyperparameters which we determined in preliminary experiments. We found the following set of parameters to perform well for both datasets: patch size $p = 256$  pixels, down scale factor $d=2$, learning rate $\eta = 10^{-3}$ and a regularization parameter $\lambda=50$. We used a stride $s=60$ pixels for the CRC dataset and a stride of $s=50$ pixels for the CTCL dataset. The smaller stride for the latter effectively led to a larger amount of training patches preventing problems with overfitting.

\subsection{Evaluation}

After training, we use the set of all images with patient outcome group $c=0$, denoted as $\mathcal{X}^{(0)}$, to generate the set of counterfactual images $\mathcal{Y}^{(1)} $, by inputting each $x^{(0)}$ into our trained generator function $G_\theta$, i.e. $\mathcal{Y}^{(1)} = \{ G_\theta(x^{(0)}) \ \  \forall \ \ x^{(0)} \in X^{(0)} \} $. Comparing the images from $\mathcal{X}^{(0)}$ with their counterfactual counterparts $\mathcal{Y}^{(1)}$, allows visual exploration of the results. We pay particular attention to the positive and negative components of the difference map $G_\theta(x^{(0)}) - x^{(0)}$ between generated and input images, which can give us insights into which proteins need to be expressed more or less in order to make the sample change the patient outcome group and in which parts of the cell structure those changes need to occur.   

To further explore differences of protein expression of different CODEX channels on a population level, we introduce three strategies for deriving quantitative measures. 

\subsubsection{Mean channel variation}

First, we investigate the mean channel variation ($\text{MCV}_k$), which measures how much each channel $k$ changes after transformation with $G_\theta$ when averaged over all images $x^{(0)}$ and the spatial dimensions $i,j$, i.e. 
\begin{equation}
\label{eq:mcv}
    \text{MCV}_k = \frac{1}{Z} \sum_{x^{(0)} \in \mathcal{X}^{(0)}} \sum_{ij} \left[G_\theta(x^{(0)}) - x^{(0)}\right]_{ijk}.
\end{equation}
Here the first sum is over all images $x^{(0)}$ from patient outcome group $c=0$. $Z$ is a normalisation factor which ensures that the largest per-channel change (negative or positive) is one. This gives us an indication of how each CODEX channel changes on average for all images of a certain outcome group. 

\subsubsection{Absolute channel variation}

Note that not all changes that can occur are per channel additions or per channel subtractions. An interesting category of changes occurs when pixel intensities are added to some areas of an image and subtracted from other areas of the image. In these cases the positive and negative contributions would cancel each other out in the computation of $\text{MCV}_k$. To detect these changes, we additionally introduce the absolute channel variation ($\text{APC}_k$) measure defined as 
\begin{equation}
\label{eq:apc}
    \text{ACV}_k = \frac{1}{Z} \sum_{x^{(0)} \in \mathcal{X}^{(0)}} \sum_{ij} \left[\vert G_{\theta}(x^{(0)}) - x^{(0)} \vert \right]_{ijk}.
\end{equation}
It differs from $\text{MCV}_k$ only by the absolute value inside the square brackets. $Z$ is again a normalisation factor scaling the per-channel measures such that the largest absolute channel variation is one. Note that $\text{ACV}_k > 0$ when the generator function $G_\theta$ adds and subtracts pixel intensities in equal degrees in different image regions of a channel $k$.

\subsubsection{Identifying statistically significant CODEX channel differences between outcome groups}
\label{sec:statistical-tests}

We also explore our method's ability to identify statistically significant differences in the protein expression of different CODEX channels for different patient outcome groups. To this end, we apply statistical significance tests on the average protein expression per image for different patient outcome groups, with the null hypothesis that they do not differ. Rejection of this null hypothesis allows us to identify CODEX channels with significantly different protein expression for two patient outcome groups. 

In addition to the sets $\mathcal{X}^{(0)}$, $\mathcal{Y}^{(1)}$ of real input images of class $c=0$, and generated counterfactual images of class $c=1$, we additionally define the set of all real images of class $c=1$ as $\mathcal{X}^{(1)}$. From these image sets we further derive the per-channel average pixel values $\mathcal{A}_k^{(0)}$, $\mathcal{A}_k^{(1)}$, $\mathcal{B}_k^{(1)}$, by averaging over the spatial dimensions for each channel of each image from $\mathcal{X}^{(0)}$, $\mathcal{X}^{(1)}$, and $\mathcal{Y}^{(1)}$, respectively. This means, the sets $\mathcal{A}_k^{(0)}$, $\mathcal{A}_k^{(1)}$, and $\mathcal{B}_k^{(1)}$ are a collection of scalars of average pixel values, one for for each input image. 

We examine two strategies for identifying CODEX channels that differ statistically significantly between two patient groups. The naive baseline strategy is to apply an unpaired Student's $t$-test to the samples $\mathcal{A}_k^{(0)}$, $\mathcal{A}_k^{(1)}$ derived from real images from both outcome groups. Note that this strategy does not involve our CF-HistoGAN framework and can be performed using the image samples alone. We hypothesise that, since the samples are \textit{unpaired}, the sensitivity of such a test will be limited, due to the fact that image samples from the two groups are structurally substantially different and the average CODEX channel values are dominated by those structural variations. The second strategy uses samples generated using the CF-HistoGAN framework to perform a \textit{paired} Student's $t$-test. Specifically, we compare the values $\mathcal{A}_k^{(0)}$, and $\mathcal{B}_k^{(1)}$ derived from the real images $\mathcal{X}^{(0)}$ of patient group $c=0$, and the generated images $\mathcal{Y}^{(1)}$ which differ only in the attributes related to the patient group, but are otherwise structurally highly similar. Therefore, we further hypothesise that we will be able to identify differences in protein expression between patient outcome groups with much higher significance. 

We report the $p$-values for both the unpaired statistical tests, and the ``artificially paired'' tests for all experiments in this paper. 

\section{Results}

In this section, we describe the results of applying our proposed CF-HistoGAN to the CRC and CTCL datasets outlined earlier (see Section~\ref{sec:material-and-equipment}). For the CRC dataset, we transformed the images from CLR (good survival) to DII (poor survival) and vice versa. For the CTCL dataset we transformed the images from the treatment responders to the non-responders and vice versa. For each of these four cases, we trained an instance of CF-HistoGAN and obtained predictions for the entire dataset as described in Section~\ref{sec:training}. The results for each dataset and each direction of transformation are shown in Figs.~\ref{fig:CTD}-\ref{fig:NTR}. For each transformation direction we show example images of the protein channels which were changed the most by our counterfactual image generator. Specifically, in each case we show the images before the transformation, the subtracted pixels, the added pixels, and the output after the transformation (see Figs.~\ref{fig:CTD}-\ref{fig:NTR}, A). The seven channels with the most changes were determined separately for each transformation direction by calculating the absolute channel variations per channel using Equation~\ref{eq:apc}. The mean and absolute channel variation values $\text{MCV}_k$ and $\text{ACV}_k$ for those seven channels are shown in panel B for each experiment, with blue and orange bars, respectively see (Figs. \ref{fig:CTD}-\ref{fig:NTR}, B). The results of the unpaired, and ``artificially paired'' comparison strategies (detailed in Section~\ref{sec:statistical-tests}) for the seven most changed channels are shown in panel C of Figs.~\ref{fig:CTD}-\ref{fig:NTR}. Lower $p$-values indicate more significantly different CODEX channel adsorptions between two patient groups. Note that we only expect significance differences for protein markers that in fact different for the two patient outcome groups. Lastly, for each figure the H\&E-stained input images for a representative TMA core (Figs. \ref{fig:CTD}-\ref{fig:NTR}, D) and composite images of the seven channels with the highest absolute channel variation before and after transformation using CF-HistoGAN are depicted (Figs. \ref{fig:CTD}-\ref{fig:NTR}, E and F). For all experiments we also report the mean and absolute channel variation values $\text{MCV}_k$ and $\text{ACV}_k$, as well as the unpaired, and ``artificially paired'' $p$-values in Tables~1-4 at the end of the document. In the following sections we describe our findings in detail.

\subsection{Transformation of CLR to DII in the colorectal cancer dataset}

\begin{figure}
    \centering
    \includegraphics[width=0.99\textwidth]{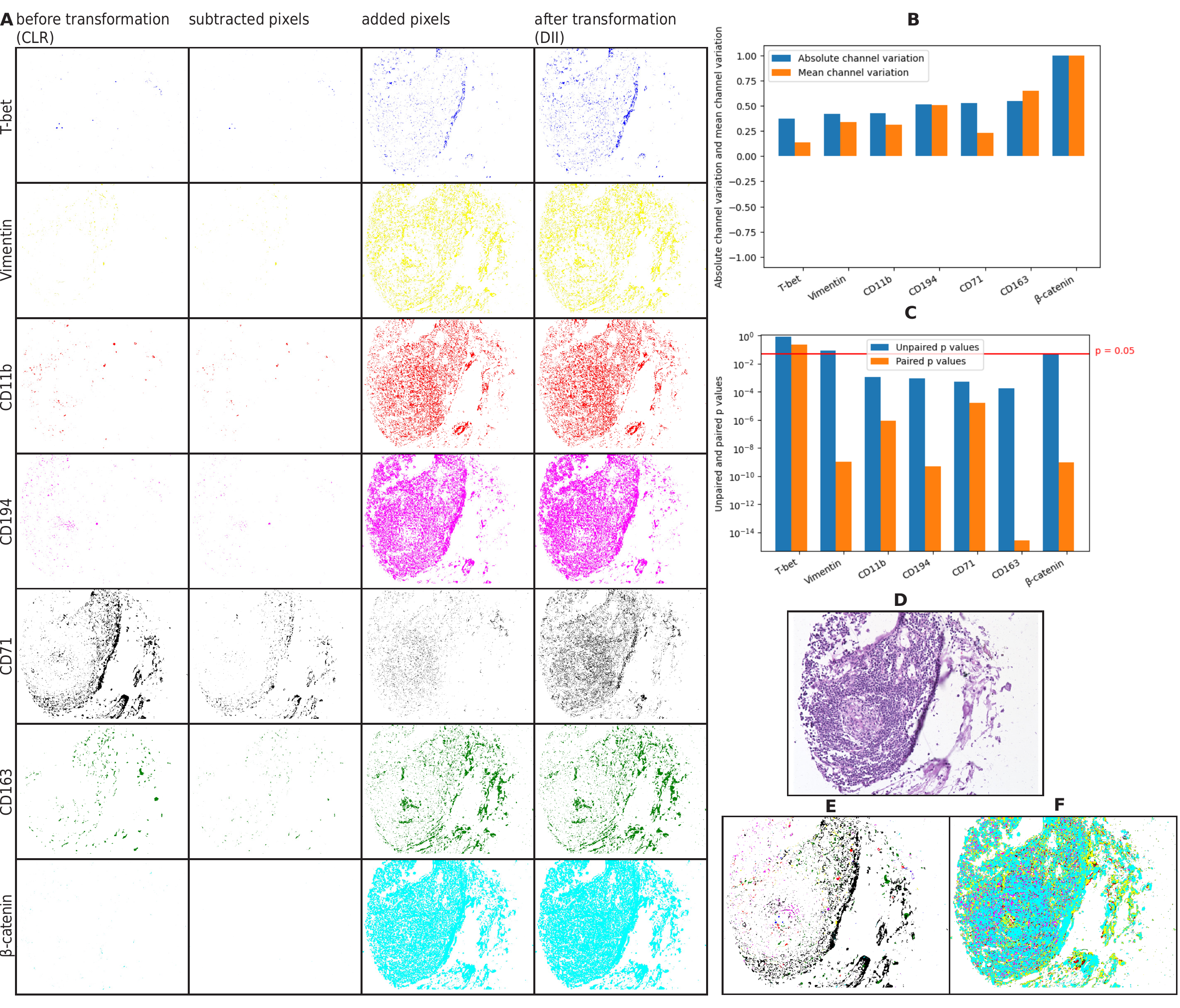}
    \caption{Artificial transformation of CLR to DII CRC samples in silico. The seven most changed channels across patients are shown. (A) Representative images of a tissue core from a CLR patient that was transformed to the DII group, depicted as images before transformation, added and subtracted pixels, and after transformation. (B) Metrics for the absolute channel variation ($\text{ACV}_k$) and mean channel variation ($\text{MCV}_k$) for each channel for the seven most changed channels across all tissue cores. (C) Unpaired $t$-test that determines if the differences in pixel values per channel between the original images of both classes (CLR vs. DII) are statistically significant (blue bars). This test is compared to a paired $t$-test (orange bars) that examines the pixel intensities in the channels of the original vs. the artificially generated images (CLR vs. CLR transformed to DII). (D) H\&E of the representative image sample (untransformed). (E-F) Composite image of the seven most changed channels shown in panel A before transformation (E) and after transformation (F).}
    \label{fig:CTD}
\end{figure}

The results of transforming TMA cores from CLR patients (good outcomes) to DII patients (bad outcomes) are shown in Figure~\ref{fig:CTD}. A representative example of the seven most changed channels across all patients is shown for one tissue core (Fig. \ref{fig:CTD}A). In all images for the seven most changed channels, pixels were added in the transformation process across all samples, as indicated by a positive value in the mean change per channel metric (Fig. \ref{fig:CTD}B, orange bars). Since this metric does not discriminate on the specific location within the tissue in which pixels are added or subtracted, we also performed an analysis of the absolute channel variation. This revealed that pixels were added in one image location and removed in another for the channels CD71 and T-bet. This is apparent through a substantially higher value in the absolute change per pixel metric in these channels than their mean change (Fig. \ref{fig:CTD}B, T-bet and CD71, blue vs. orange bars). 

When comparing the artificial tissue samples created by CF-HistoGAN to their original counterparts using a paired Student's $t$-test, we found that vimentin (a mesenchymal marker), CD11b (a marker for monocytes / macrophages), CD194 (C-C chemokine receptor 4 [CCR4]), CD71 (transferrin receptor), CD163 (a marker for immunosuppressive M2 macrophages) and $\beta$-catenin (a signaling protein and potential oncogene) were significantly increased after transformation from CLR to DII (Fig. \ref{fig:CTD}C). The $p$-values for all seven channels were lower (i.e. more significant) for the paired $t$-test compared to the unpaired $t$-test (Fig. \ref{fig:CTD}C, blue vs. orange bars). The changes in vimentin did not reach statistical significance (threshold of $p = 0.05$) in the unpaired $t$-tests, and their differences were only revealed in the paired $t$-test after transformation by CF-HistoGAN (Fig. \ref{fig:CTD}C). 

\subsection{Transformation of DII to CLR in the colorectal cancer dataset}

\begin{figure}
    \centering
    \includegraphics[width=0.99\textwidth]{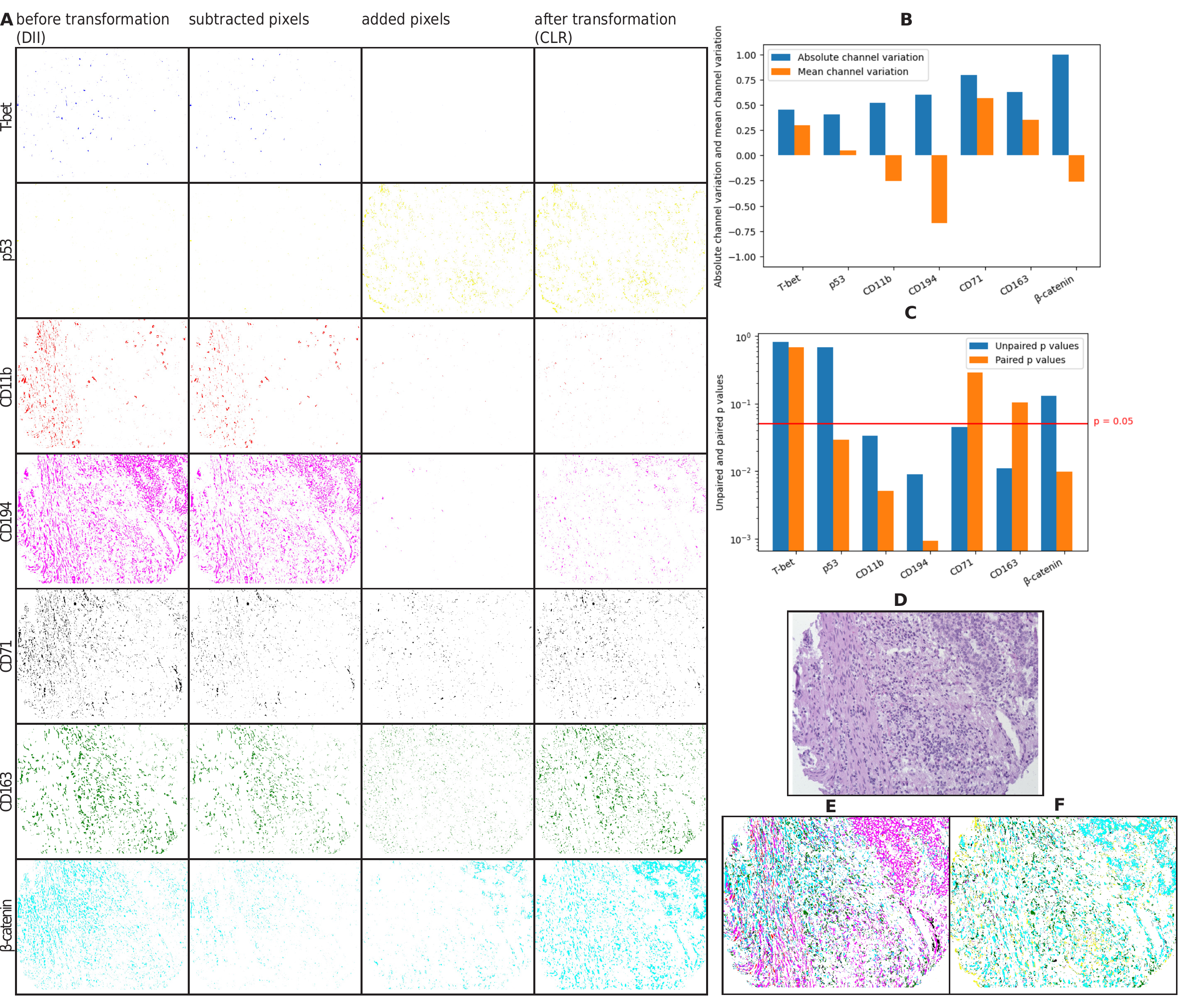}
    \caption{Artificial transformation of CRC DII to CLR samples in silico. The seven most changed channels across patients are shown. (A) Representative images of a tissue core from a DII patient that was transformed to the CLR group. (B) Metrics for the absolute channel variation ($\text{ACV}_k$) and mean channel variation ($\text{MCV}_k$). (C) Unpaired (blue bars) and paired (orange bars) $t$-test between images of both patient outcome groups. (D) H\&E of the representative image sample (untransformed). (E-F) Composite image of the seven most changed channels shown in panel A before transformation (E) and after transformation (F).}
    \label{fig:DTC}
\end{figure}

Figure \ref{fig:DTC} illustrates the transformation process of images of TMA cores from DII patients (bad outcomes) to resemble those of CLR patients (good outcomes). A representative example of the seven most changed channels across all patients is shown for one tissue core (Fig. \ref{fig:DTC}A). Pixel values were predominantly added for T-bet, CD163 and CD71 (indicated by a positive value in the mean change metric), while for CD194, CD11b and $\beta$-catenin pixels were predominantly subtracted (indicated by a negative value in the mean change metric) (Fig. \ref{fig:DTC}B, orange bars). For p53, CD11b, CD194, CD163 and $\beta$-catenin, the absolute change per pixel metric was higher than the mean change metric, indicating simultaneous removal and addition of pixel intensities in these channels (Fig. \ref{fig:DTC}B). This “shift” of pixels to different positions in the channel image is clearly visible for $\beta$-catenin (Fig. \ref{fig:DTC}A, top row, before vs. after transformation).

When comparing the artificial tissue samples created by CF-HistoGAN to their original counterparts using a paired $t$-test, we found that p53, CD11b, CD194 and $\beta$-catenin were significantly increased after transformation from DII to CLR (Fig. \ref{fig:DTC}C). Moreover, these changes were more statistically significant than the differences between the original images of both classes for these markers, since the paired $t$-test $p$-values were lower than those from the unpaired $t$-tests, while for CD163 and CD71 the changes were less statistically significant than the differences between the original images of both classes, since the paired $t$-test $p$-values were $> 0.05$ (Fig. \ref{fig:DTC}C, blue vs. orange bars).

\subsection{Transformation of therapy responders to non-responders in the cutaneous T cell lymphoma dataset}

\begin{figure}
    \centering
    \includegraphics[width=0.99\textwidth]{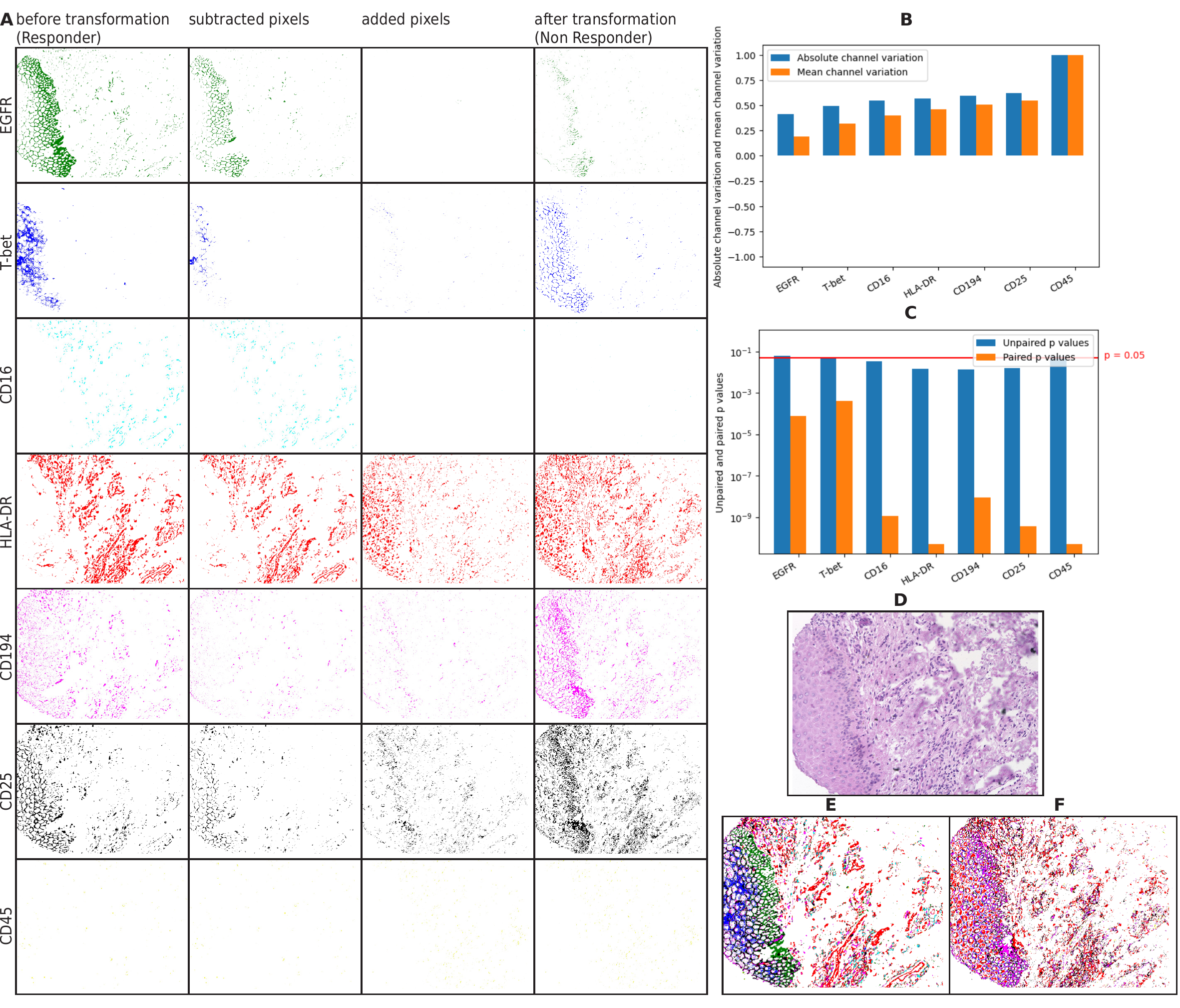}
    \caption{Artificial transformation of CTCL responder to non-responder samples in silico. The seven most changed channels across patients are shown. (A) Representative images of a tissue core from a DII patient that was transformed to the CLR group. (B) Metrics for the absolute channel variation ($\text{ACV}_k$) and mean channel variation ($\text{MCV}_k$). (C) Unpaired (blue bars) and paired (orange bars) $t$-test between images of both patient outcome groups. (D) H\&E of the representative image sample (untransformed). (E-F) Composite image of the seven most changed channels shown in panel A before transformation (E) and after transformation (F).}
    \label{fig:RTN}
\end{figure}

Figure \ref{fig:RTN} illustrates the transformation process from images of TMA cores of responder patients to resemble those of non-responder patients. A representative example for the seven most changed channels of one tissue core is shown in Fig. \ref{fig:RTN}A. In all images for the seven most changed channels, pixels were added to the channels in the transformation process across all samples as indicated by a positive value in the mean change metric (Fig. \ref{fig:RTN}B). For EGFR, T-bet and CD16 the absolute change per pixel metric was substantially higher than the mean change metric, indicating that again, for those channels pixel intensities were simultaneously added and removed in different locations of a single image (Figs. \ref{fig:RTN}B). For instance, such “shifts” can be observed in the epidermis of the skin (Fig. \ref{fig:RTN}A, D, E, F, left portion of the image), where pixels in the CD194 and HLA-DR channels were added, while pixels in the EGFR and T-bet channels were removed (Fig. \ref{fig:RTN}A, E, F, left). 
The changes between the artificial tissue samples created by the CF-HistoGAN compared to their original counterparts were statistically significant in the seven most changed channels, as analyzed by paired $t$-tests (Fig. \ref{fig:RTN}C, orange bars). Furthermore, these changes were more statistically significant than the differences between the original images of both classes analyzed by unpaired $t$-tests (Fig. \ref{fig:RTN}C, blue bars).

\subsection{Transformation of therapy non-responders to responders in the cutaneous T cell lymphoma dataset}

\begin{figure}
    \centering
    \includegraphics[width=0.99\textwidth]{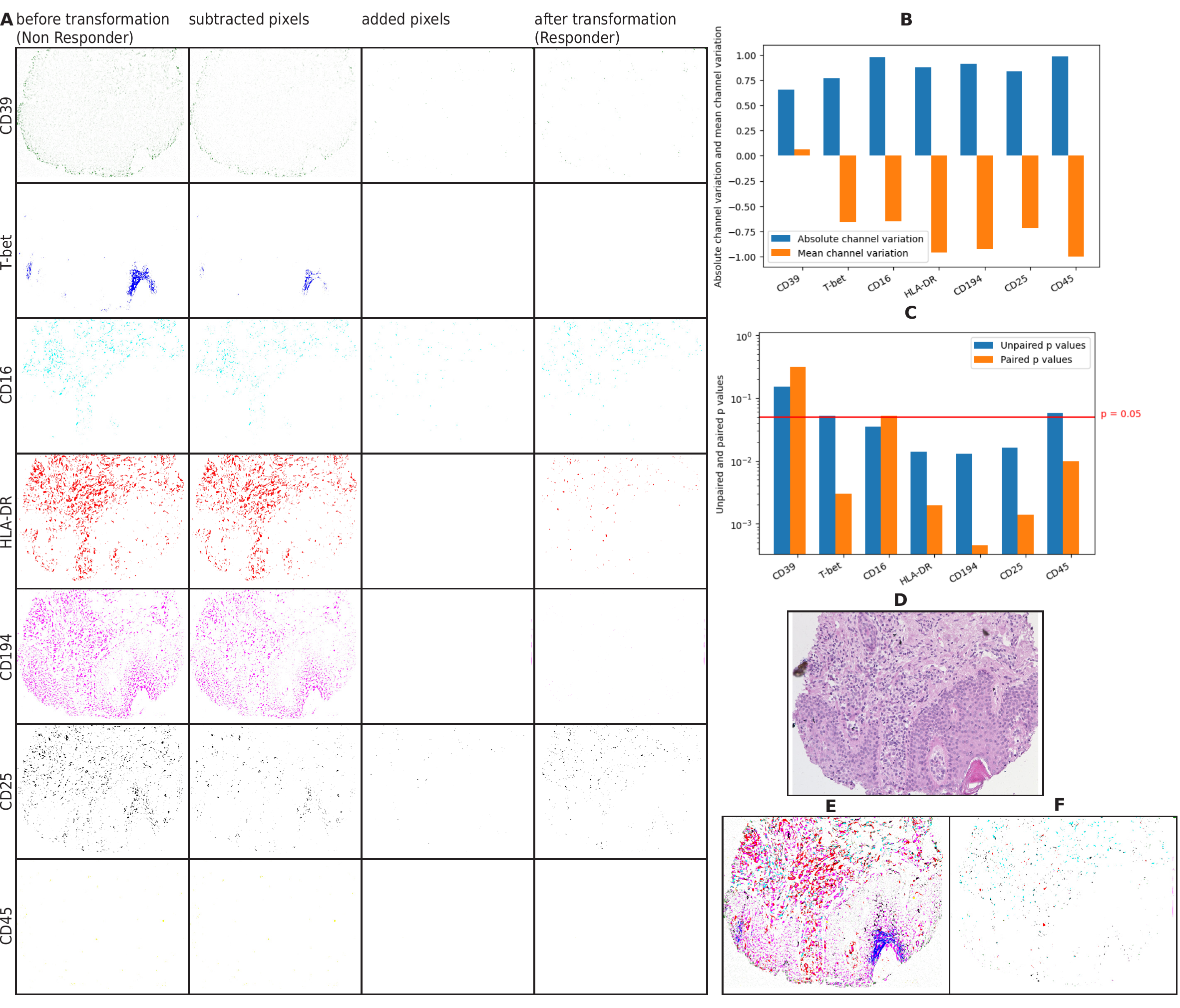}
    \caption{Artificial transformation of CTCL non-responder to responder samples in silico. The seven most changed channels across patients are shown. (A) Representative images of a tissue core from a DII patient that was transformed to the CLR group. (B) Metrics for the absolute channel variation ($\text{ACV}_k$) and mean channel variation ($\text{MCV}_k$). (C) Unpaired (blue bars) and paired (orange bars) $t$-test between images of both patient outcome groups. (D) H\&E of the representative image sample (untransformed). (E-F) Composite image of the seven most changed channels shown in panel A before transformation (E) and after transformation (F).}
    \label{fig:NTR}
\end{figure}

Figure \ref{fig:NTR} illustrates the transformation process from images of TMA cores of treatment non-responder patients to resemble those of responders. A representative example for the seven most changed channels of one tissue core is shown in Fig. \ref{fig:NTR}A. In all images for the seven most changed channels, pixels were subtracted in the transformation process across all samples as indicated by a negative value in the mean change metric, except for CD39 in which the mean change per channel was near 0 (Fig. \ref{fig:NTR}B).

For CD39 and CD16 the absolute channel variation per channel is higher than the mean change per channel, which indicates that, as previously observed, for those channels pixel intensities were added an removed to a similar degree in different regions of individual images. 

The changes between the artificial tissue samples created by the CF-HistoGAN to their original counterparts are statistically significant in T-bet, CD25, HLA-DR, CD194 and CD45 (Fig. \ref{fig:NTR}C). Furthermore, these changes are more statistically significant than the differences between the original images of both classes, as the $p$-value in the paired $t$-test between the original images and their counterfactuals is lower than the $p$-value in the unpaired $t$-test between the original images of both classes, while for CD39 and CD16 the changes were less statistically significant than the differences between the original images of both classes (Fig. \ref{fig:NTR}C, blue vs. orange bars).

\section{Discussion}

Here, we present CF-HistoGAN, a machine learning framework that creates artificial paired samples in HMTI datasets to uncover biologically relevant signatures in tissue protein expression. We applied CF-HistoGAN to CODEX HMTI datasets to transform cancer patient tissue images between different groups (bad outcomes into good outcomes; therapy non-responders into responders; and vice versa). Interestingly, we observed significant changes in the amount of pixel intensities in different protein channels. In the CRC dataset, when transforming CLR to DII images, CD163 showed the most significant changes in the paired $t$-test, followed by CD71 and CD194. Since CD163 is a marker for immunosuppressive M2 macrophages and higher expression in the iTME is associated with worse outcomes in CRC~\citep{herrera_cancer-associated_2013}, the increased amount of CD163 in CLR images transformed to the DII group (i.e., from good to bad outcomes) was not surprising. However, we did not observe the opposite in images transformed from DII to CLR, i.e., CD163 was not subtracted here. However, the paired $t$-test $p$-value for this direction of transformation was not significant. CD11b, a monocyte marker, behaved in a similar fashion. The transferrin receptor CD71 was also added to the samples when transforming CLR to DII. CD71 has been shown to be upregulated in CRC compared to normal colon mucosa~\citep{prutki_altered_2006}, and increased CD71 expression is associated with poor prognosis and therapy resistance in breast cancer~\citep{habashy_transferrin_2010}. Moreover, $\beta$-catenin, a key molecule in the WNT signaling pathway that is important in CRC oncogenesis~\citep{brabletz_-catenin_2002}, was significantly increased after transformation from CLR to DII. Here, the significance of this increase could only be detected by paired $t$-test, since the comparison of untransformed images using an unpaired $t$-test did not reveal statistically significant differences. Transformation in the opposite direction (i.e., DII to CLR) resulted in a removal of $\beta$-catenin pixels as indicated by the mean change metric.

In CTCL, we observed interesting changes. CD25, the interleukin 2 receptor $\alpha$ chain, was significantly increased after transformation from responders to non-responders. Vice versa, it was significantly reduced after transformation from non-responders to responders. CD25 is an important survival factor for T cells, is up-regulated on CTCL tumor cells (which are CD4+ T cells), and higher CD25 expression levels are associated with higher CTCL tumor stage and histological grade~\citep{talpur_cd25_2006,phillips_immune_2021}. Similar changes were observed for CD45, a pan-leukocyte marker expressed on all T cells, suggesting higher tumor cell load after transformation from responders to non-responders and lower tumor cell load after transformation in the opposite direction. We also observed significant changes in the levels of HLA-DR (human leukocyte antigen DR, also known as major histocompatibility complex class [MHC-II]) in both directions, with HLA-DR levels higher in non-responders  and lower in responders after transformation. HLA-DR / MHC-II expression in the iTME plays an important role in antitumoral immunity (Axelrod et al., 2019); however, the significance of our findings for this molecule are somewhat counterintuitive and less clear.

One of the most interesting and consistent findings we observed was the change in levels of CD194 (CCR4), which was present in both patient groups / disease types in a similar manner. In CRC, CD194 was increased after transformation from CLR to DII and reduced after transformation in the opposite direction. Likewise, in CTCL, CD194 was increased after transformation from responders to non-responders and reduced after transformation from non-responders to responders. CD194 / CCR4 is a chemokine receptor that is predominantly expressed by T cells, and it has two ligands, C-C chemokine ligand 17 (CCL17) and CCL22. CCL22 expression was shown to be increased in CRC, and both CCL22 and CCL17 can recruit CCR4-expressing immunosuppressive regulatory T cells into the iTME~\cite{korbecki2020cc}. CCR4 is highly expressed in CTCL, which has led to the development of the anti-CCR4 monoclonal antibody mogamulizumab, which is an approved immunotherapy for the treatment of CTCL~\citep{kim2018mogamulizumab,yoshie2021ccr4}. 
In summary, these findings show that CF-HistoGAN can uncover biologically significant changes in the expression of molecules in HMTI data, which could lead to the discovery of novel biomarkers and/or therapeutic targets.

Our method performed well in finding the most important biomarkers and their role on the outcome of an immunotherapy. E.g., our method found out that the chemokine receptor CD194 is associated with a bad therapy outcome in CRC and CTCL. However, the method did not perform consistently across all networks, with the transformation from CLR to DII (CRC) and responder to non-responder (CTCL) showing better results than the reverse transformations of DII to CLR (CRC) and non-responder to responder (CTCL). This may be due to an imbalance of samples in the groups or the presence of anatomical structures such as follicles within the images, which makes it easier for the network to determine where to add or subtract pixels.

The method offers opportunities for improvement and optimization. CF-HistoGAN operates directly on raw pixels which makes it highly flexible. However, it does not take advantage of additional domain knowledge that we possess. For instance, our method does not leverage any structural priors. The performance of CF-HistoGAN may be improved in the future by also including cell segmentation maps as an additional input which may help to further guide the generation process. Moreover, the method does not use the information of individual cell types which can in principle be derived from the CODEX data. An approach operating on explicitly expressed cell neighborhoods may reduce the problems complexity and thus perform more robustly. Moreover, generated counterfactual cell neighbourhoods may be better suited to understand the iTME and generate biological hypotheses. 

\begin{table}[h]
\begin{tabular}{lllll}
\textbf{Channel Name} & $\text{ACV}_k$ & $\text{MCV}_k$ & \textbf{Unpaired p-values} & \textbf{Paired p-values} \\
beta-catenin   & 1      & 1        & 0.0457   & 9.62e-10 \\
CD163          & 0.548  & 0.651    & 0.000168 & 2.59e-15 \\
CD71           & 0.529  & 0.231    & 0.000496 & 1.55e-05 \\
CD194          & 0.518  & 0.505    & 0.000854 & 5.04e-10 \\
CD11b          & 0.428  & 0.313    & 0.00111  & 8.56e-07 \\
Vimentin       & 0.421  & 0.338    & 0.0789   & 1.02e-09 \\
T-bet          & 0.374  & 0.136    & 0.797    & 0.218    \\
CD44           & 0.364  & 0.186    & 0.811    & 0.00154  \\
aSMA           & 0.343  & 0.145    & 0.0493   & 0.0245   \\
CDX2           & 0.33   & 0.323    & 0.599    & 1.11e-05 \\
Podoplanin     & 0.323  & 0.229    & 0.748    & 1.37e-07 \\
CD34           & 0.312  & 0.222    & 0.793    & 3.55e-10 \\
CD3            & 0.301  & 0.117    & 0.663    & 1.38e-06 \\
CD68           & 0.297  & 0.274    & 3.36e-05 & 1.1e-11  \\
CD45RO         & 0.297  & 0.129    & 0.02     & 0.000377 \\
p53            & 0.284  & -0.153   & 0.768    & 0.00467  \\
CD5            & 0.255  & 0.181    & 0.0479   & 6.2e-05  \\
PD-1           & 0.253  & 0.184    & 0.446    & 3.06e-05 \\
IDO-1          & 0.252  & 0.161    & 0.168    & 0.000177 \\
CD25           & 0.246  & 0.061    & 0.449    & 0.105    \\
EGFR           & 0.237  & 0.182    & 0.00898  & 2.91e-06 \\
CD4            & 0.214  & 0.168    & 0.118    & 1.09e-06 \\
FOXP3          & 0.212  & -0.0223  & 0.0774   & 0.274    \\
CD8            & 0.209  & 0.135    & 0.0104   & 6.47e-10 \\
Collagen IV    & 0.204  & 0.11     & 0.266    & 0.000272 \\
VISTA          & 0.195  & -0.0153  & 0.109    & 0.461    \\
MMP12          & 0.195  & 0.012    & 0.779    & 0.53     \\
CD56           & 0.188  & 0.15     & 0.0606   & 8.08e-05 \\
ICOS           & 0.188  & 0.0193   & 0.703    & 0.224    \\
Na-K-ATPase    & 0.18   & 0.1      & 0.305    & 6.08e-06 \\
CD2            & 0.173  & 0.00524  & 0.222    & 0.971    \\
CD21           & 0.171  & 0.0722   & 0.418    & 0.000307 \\
CD30           & 0.17   & 0.0615   & 0.769    & 0.0403   \\
GATA3          & 0.157  & 0.0331   & 0.91     & 0.371    \\
CD57           & 0.147  & -0.0245  & 0.0478   & 0.207    \\
LAG-3          & 0.141  & 0.0215   & 0.462    & 0.432    \\
GFAP           & 0.141  & 0.0976   & 0.923    & 0.00164  \\
CD31           & 0.137  & 0.146    & 0.179    & 1.48e-09 \\
CD45           & 0.132  & -0.0158  & 0.161    & 0.401    \\
BCL-2          & 0.125  & 0.0236   & 0.34     & 0.318    \\
CD38           & 0.117  & 0.0435   & 0.658    & 0.023    \\
CD11c          & 0.116  & 0.0611   & 0.26     & 0.00219  \\
MUC-1          & 0.1    & 0.000723 & 0.288    & 0.663    \\
Ki67           & 0.0994 & 0.0205   & 0.092    & 0.00774  \\
MMP9           & 0.0932 & -0.0834  & 0.0116   & 1.51e-05 \\
PD-L1          & 0.0899 & 0.0342   & 0.24     & 0.00513  \\
CD45RA         & 0.0846 & -0.005   & 0.236    & 0.712    \\
Synaptophysin  & 0.0826 & -0.023   & 0.459    & 0.154    \\
Chromogranin A & 0.0696 & -0.0426  & 0.162    & 0.00323  \\
Cytokeratin    & 0.0691 & 0.0106   & 0.915    & 0.585    \\
HLA-DR         & 0.0574 & -0.00335 & 0.135    & 0.603    \\
CD7            & 0.0503 & 0.000311 & 0.623    & 0.76     \\
CD20           & 0.0465 & -0.0125  & 0.0244   & 0.287    \\
CD15           & 0.0459 & 0.00286  & 0.0501   & 0.794    \\
CD138          & 0.0382 & -0.00131 & 0.794    & 0.385    \\
Granzyme B     & 0.024  & -0.0154  & 0.0518   & 0.00279 
\end{tabular}
\label{tab:CTD}
\caption{All quantitative results for CRC transforming images from CLR to DII.}
\end{table}

\begin{table}[h]
\begin{tabular}{lllll}
\textbf{Channel Name} & $\text{ACV}_k$ & $\text{MCV}_k$ & \textbf{Unpaired p-values} & \textbf{Paired p-values} \\
beta-catenin   & 1      & -0.264   & 0.13    & 0.00992  \\
CD71           & 0.799  & 0.567    & 0.0455  & 0.287    \\
CD163          & 0.632  & 0.356    & 0.0111  & 0.105    \\
CD194          & 0.604  & -0.67    & 0.00895 & 0.000943 \\
CD11b          & 0.523  & -0.255   & 0.0332  & 0.00513  \\
T-bet          & 0.454  & 0.296    & 0.825   & 0.683    \\
p53            & 0.408  & 0.0492   & 0.689   & 0.0293   \\
CD44           & 0.402  & 1        & 0.953   & 0.272    \\
aSMA           & 0.385  & -0.817   & 0.182   & 8.29e-06 \\
CD45RO         & 0.333  & -0.559   & 0.159   & 0.00011  \\
Vimentin       & 0.333  & -0.649   & 0.25    & 1.11e-06 \\
CD68           & 0.302  & -0.719   & 0.00783 & 1.19e-06 \\
CD3            & 0.301  & -0.0561  & 0.928   & 0.0196   \\
CDX2           & 0.299  & -0.0625  & 0.274   & 0.2      \\
CD5            & 0.298  & -0.208   & 0.124   & 0.00582  \\
CD25           & 0.293  & 0.353    & 0.678   & 0.138    \\
Podoplanin     & 0.28   & -0.067   & 0.966   & 0.679    \\
EGFR           & 0.276  & -0.33    & 0.0661  & 0.000322 \\
IDO-1          & 0.273  & -0.00776 & 0.592   & 0.0546   \\
FOXP3          & 0.271  & 0.492    & 0.409   & 0.162    \\
VISTA          & 0.257  & 0.376    & 0.374   & 0.0179   \\
PD-1           & 0.249  & -0.0166  & 0.625   & 0.113    \\
MMP12          & 0.219  & 0.0549   & 0.767   & 0.899    \\
ICOS           & 0.213  & 0.0451   & 0.907   & 0.872    \\
CD56           & 0.21   & -0.0406  & 0.119   & 0.0716   \\
CD4            & 0.206  & -0.146   & 0.134   & 0.00058  \\
CD2            & 0.198  & 0.172    & 0.534   & 0.0276   \\
CD21           & 0.193  & 0.0139   & 0.51    & 0.584    \\
CD8            & 0.191  & -0.273   & 0.0798  & 8.67e-07 \\
GATA3          & 0.187  & -0.0348  & 0.863   & 0.713    \\
Collagen IV    & 0.18   & -0.28    & 0.491   & 0.00052  \\
CD34           & 0.176  & -0.251   & 0.632   & 0.000224 \\
CD57           & 0.167  & 0.14     & 0.212   & 0.00897  \\
CD45           & 0.163  & 0.16     & 0.161   & 0.591    \\
CD30           & 0.161  & -0.0956  & 0.704   & 0.117    \\
LAG-3          & 0.154  & 0.0764   & 0.58    & 0.588    \\
Na-K-ATPase    & 0.153  & -0.188   & 0.549   & 0.00197  \\
CD45RA         & 0.14   & 0.29     & 0.321   & 0.647    \\
CD38           & 0.135  & 0.0682   & 0.566   & 0.491    \\
BCL-2          & 0.135  & 0.0364   & 0.499   & 0.2      \\
MUC-1          & 0.13   & -0.241   & 0.805   & 0.321    \\
CD11c          & 0.129  & -0.0776  & 0.177   & 0.0316   \\
GFAP           & 0.116  & -0.135   & 0.73    & 0.0548   \\
MMP9           & 0.107  & 0.0455   & 0.0538  & 0.855    \\
Ki67           & 0.103  & -0.131   & 0.449   & 0.00337  \\
PD-L1          & 0.0998 & -0.00187 & 0.323   & 0.0201   \\
Synaptophysin  & 0.0928 & -0.0853  & 0.454   & 0.143    \\
Chromogranin A & 0.0797 & -0.0419  & 0.277   & 0.226    \\
Cytokeratin    & 0.069  & -0.0799  & 0.824   & 0.117    \\
CD31           & 0.069  & -0.149   & 0.322   & 8.85e-06 \\
CD15           & 0.0668 & -0.113   & 0.02    & 0.000259 \\
CD20           & 0.0656 & 0.105    & 0.0367  & 0.0657   \\
CD7            & 0.0558 & -0.00882 & 0.922   & 0.00931  \\
HLA-DR         & 0.0546 & 0.0158   & 0.143   & 0.785    \\
CD138          & 0.0418 & -0.0316  & 0.849   & 0.056    \\
Granzyme B     & 0.0297 & 0.0146   & 0.543   & 0.657   
\end{tabular}
\label{tab:DTC}
\caption{All quantitative results for CRC transforming images from DII to CLR.}
\end{table}

\begin{table}[h]
\begin{tabular}{lllll}
\textbf{Channel Name} & $\text{ACV}_k$ & $\text{MCV}_k$ & \textbf{Unpaired p-values} & \textbf{Paired p-values} \\
CD45              & 1      & 1        & 0.0576  & 5.2e-11  \\
CD25              & 0.62   & 0.548    & 0.0161  & 3.54e-10 \\
CD194             & 0.599  & 0.51     & 0.0131  & 9.14e-09 \\
HLA-DR            & 0.567  & 0.461    & 0.0141  & 5.11e-11 \\
CD16              & 0.547  & 0.403    & 0.035   & 1.18e-09 \\
T-bet             & 0.494  & 0.319    & 0.0516  & 0.000425 \\
EGFR              & 0.412  & 0.193    & 0.0627  & 7.7e-05  \\
CD164             & 0.408  & 0.356    & 0.0017  & 3.48e-08 \\
CD3               & 0.343  & 0.2      & 0.188   & 0.000689 \\
CD4               & 0.314  & 0.222    & 0.105   & 5.42e-06 \\
CD7               & 0.311  & 0.0745   & 0.864   & 0.334    \\
CD39              & 0.311  & 0.0205   & 0.151   & 0.318    \\
CD71              & 0.31   & 0.102    & 0.0118  & 2.72e-05 \\
CCR6              & 0.285  & 0.175    & 0.048   & 3.11e-05 \\
CD34              & 0.277  & 0.0231   & 0.262   & 0.997    \\
CD138             & 0.258  & 0.0833   & 0.921   & 0.121    \\
CD68              & 0.257  & 0.158    & 0.0372  & 1.13e-07 \\
Vimentin          & 0.254  & 0.0782   & 0.0963  & 0.00679  \\
CD11b             & 0.25   & 0.0772   & 0.0297  & 0.00223  \\
MUC-1             & 0.247  & -0.0725  & 0.11    & 0.0334   \\
beta-catenin         & 0.247  & 0.0669   & 0.48    & 0.165    \\
CD5               & 0.233  & 0.166    & 0.0181  & 1.08e-07 \\
BCL-2             & 0.217  & 0.191    & 0.0104  & 2.8e-09  \\
p53               & 0.214  & 0.11     & 0.0103  & 0.00156  \\
CLA-CD162         & 0.213  & 0.136    & 0.00813 & 7.79e-08 \\
CollagenIV        & 0.211  & 0.0407   & 0.928   & 0.00735  \\
CD45RA            & 0.21   & 0.204    & 0.439   & 3.19e-09 \\
MMP12             & 0.209  & 0.0845   & 0.0246  & 0.00156  \\
CD45RO            & 0.2    & 0.178    & 0.00889 & 2.73e-09 \\
CD38              & 0.195  & 0.106    & 0.0193  & 3.04e-07 \\
CD163             & 0.184  & 0.0759   & 0.0504  & 3.75e-05 \\
GATA3             & 0.167  & 0.0701   & 0.0732  & 0.00131  \\
CD69              & 0.137  & 0.0512   & 0.0247  & 0.0112   \\
Ki-67             & 0.135  & 0.0835   & 0.136   & 4.14e-09 \\
ICOS              & 0.131  & 0.0145   & 0.894   & 0.864    \\
PD-1              & 0.13   & 0.076    & 0.2     & 0.000226 \\
CD11c             & 0.119  & 0.111    & 0.0648  & 7.56e-11 \\
PanCytokeratin    & 0.117  & 0.0296   & 0.985   & 0.309    \\
PD-L1             & 0.117  & 0.0674   & 0.0114  & 3.01e-05 \\
VISTA             & 0.109  & 0.0332   & 0.139   & 0.0106   \\
Podoplanin        & 0.105  & 0.0569   & 0.0277  & 3.76e-07 \\
CD8               & 0.104  & 0.063    & 0.0284  & 4.91e-05 \\
CD30              & 0.0949 & 0.0631   & 0.0121  & 7.64e-06 \\
FOXP3             & 0.0944 & 0.0315   & 0.144   & 0.000496 \\
IDO-1             & 0.0919 & 0.0567   & 0.00299 & 1.16e-05 \\
CD31              & 0.091  & 0.0401   & 0.473   & 4.57e-06 \\
CD2               & 0.0829 & 0.0652   & 0.0371  & 4.1e-05  \\
CD62L             & 0.0757 & 0.0177   & 0.257   & 0.0374   \\
CD57              & 0.0703 & 0.0274   & 0.0479  & 0.00533  \\
CD56              & 0.0682 & 0.0275   & 0.00875 & 0.000485 \\
CD20              & 0.0642 & 0.023    & 0.616   & 0.294    \\
MMP9              & 0.0635 & 0.0135   & 0.407   & 0.412    \\
CD1a              & 0.0516 & 0.0142   & 0.388   & 0.201    \\
Mastcell-tryptase & 0.0479 & -0.00686 & 0.216   & 4.95e-05 \\
LAG3              & 0.0358 & 0.00647  & 0.564   & 0.137    \\
CD15              & 0.027  & 0.00627  & 0.589   & 6.24e-05 \\
GranzymeB         & 0.0254 & 0.00728  & 0.347   & 3.52e-06
\end{tabular}
\label{tab:RTN}
\caption{All quantitative results for CTCL transforming images from treatment responder to non-responder.}
\end{table}

\begin{table}[h]
\begin{tabular}{lllll}
\textbf{Channel Name} & $\text{ACV}_k$ & $\text{MCV}_k$ & \textbf{Unpaired p-values} & \textbf{Paired p-values} \\
CD45              & 0.985  & -1       & 0.0576  & 0.00994  \\
CD16              & 0.982  & -0.651   & 0.035   & 0.0517   \\
CD194             & 0.912  & -0.925   & 0.0131  & 0.00046  \\
HLA-DR            & 0.881  & -0.963   & 0.0141  & 0.00197  \\
CD25              & 0.842  & -0.721   & 0.0161  & 0.00138  \\
T-bet             & 0.774  & -0.656   & 0.0516  & 0.003    \\
CD39              & 0.66   & 0.0636   & 0.151   & 0.311    \\
CD164             & 0.63   & -0.573   & 0.0017  & 0.000206 \\
EGFR              & 0.619  & -0.472   & 0.0627  & 0.00024  \\
CD7               & 0.597  & 0.137    & 0.864   & 0.00482  \\
Vimentin          & 0.569  & -0.071   & 0.0963  & 0.578    \\
CD71              & 0.548  & -0.143   & 0.0118  & 0.0168   \\
CD68              & 0.496  & -0.128   & 0.0372  & 0.266    \\
CCR6              & 0.486  & -0.412   & 0.048   & 0.000472 \\
MMP12             & 0.485  & -0.152   & 0.0246  & 0.0447   \\
CD11b             & 0.484  & -0.0953  & 0.0297  & 0.121    \\
CD3               & 0.444  & -0.282   & 0.188   & 0.0087   \\
CD5               & 0.422  & -0.376   & 0.0181  & 0.00167  \\
p53               & 0.407  & -0.285   & 0.0103  & 6.48e-06 \\
MUC-1             & 0.397  & 0.045    & 0.11    & 0.6      \\
CD4               & 0.393  & -0.314   & 0.105   & 0.00729  \\
CD34              & 0.384  & -0.0135  & 0.262   & 0.98     \\
beta-catenin         & 0.351  & -0.177   & 0.48    & 0.00727  \\
CLA-CD162         & 0.347  & -0.231   & 0.00813 & 0.00509  \\
CollagenIV        & 0.345  & 0.0129   & 0.928   & 0.268    \\
CD38              & 0.32   & -0.24    & 0.0193  & 0.00188  \\
GATA3             & 0.309  & -0.187   & 0.0732  & 0.0024   \\
CD138             & 0.309  & -0.0737  & 0.921   & 0.256    \\
CD163             & 0.304  & -0.0423  & 0.0504  & 0.183    \\
BCL-2             & 0.302  & -0.279   & 0.0104  & 0.00152  \\
CD45RO            & 0.292  & -0.269   & 0.00889 & 0.0141   \\
CD69              & 0.243  & -0.119   & 0.0247  & 0.00102  \\
PD-L1             & 0.205  & -0.159   & 0.0114  & 0.00167  \\
ICOS              & 0.199  & -0.0588  & 0.894   & 0.0352   \\
Podoplanin        & 0.194  & -0.0574  & 0.0277  & 0.0592   \\
CD45RA            & 0.186  & -0.108   & 0.439   & 0.196    \\
Ki-67             & 0.177  & -0.0261  & 0.136   & 0.253    \\
VISTA             & 0.176  & -0.0636  & 0.139   & 0.0131   \\
IDO-1             & 0.174  & -0.105   & 0.00299 & 0.00464  \\
CD11c             & 0.174  & -0.146   & 0.0648  & 0.0319   \\
CD30              & 0.166  & -0.114   & 0.0121  & 0.0019   \\
CD8               & 0.16   & -0.0886  & 0.0284  & 0.00653  \\
PanCytokeratin    & 0.157  & 0.0194   & 0.985   & 0.782    \\
FOXP3             & 0.156  & -0.06    & 0.144   & 0.232    \\
PD-1              & 0.155  & -0.0735  & 0.2     & 0.0334   \\
CD31              & 0.145  & -0.0122  & 0.473   & 0.844    \\
CD62L             & 0.132  & -0.0722  & 0.257   & 0.000662 \\
CD57              & 0.13   & -0.0689  & 0.0479  & 0.00277  \\
CD56              & 0.126  & -0.0349  & 0.00875 & 0.0131   \\
MMP9              & 0.124  & -0.00407 & 0.407   & 0.633    \\
CD2               & 0.122  & -0.0702  & 0.0371  & 0.0317   \\
CD1a              & 0.0991 & -0.0356  & 0.388   & 0.126    \\
CD20              & 0.0977 & -0.0156  & 0.616   & 0.469    \\
Mastcell-tryptase & 0.0777 & 0.0253   & 0.216   & 7.6e-06  \\
LAG3              & 0.0685 & 0.0174   & 0.564   & 0.497    \\
CD15              & 0.0464 & 0.0346   & 0.589   & 1.13e-08 \\
GranzymeB         & 0.0446 & 0.0332   & 0.347   & 3.86e-11
\end{tabular}
\label{tab:NTR}
\caption{All quantitative results for CTCL transforming images from non-responder to responder.}
\end{table}

\section{Acknowledgments}
We thank Mihaly Sulyok (Department of Pathology, University Hospital Tübingen, Germany) and Manfred Claassen (Department of Internal Medicine I, University Hospital Tübingen and Department of Computer Science, University of Tübingen, Germany) for critically reading the manuscript. This work is partially funded by the Deutsche Forschungsgemeinschaft (DFG, German Research Foundation) under Germany’s Excellence Strategy – EXC number 2064/1 – Project number 390727645.

\section{Conflicts of interest}
C.M.S. is a scientific advisor to AstraZeneca plc, and is on the scientific advisory board of, has stock options in, and has received research funding from Enable Medicine Inc., all outside the current work. M.P. and C.F.B. declare no conflicts of interest.

\bibliography{bibliography}

\end{document}